\documentclass[letterpaper, 10 pt, conference]{ieeeconf}  
\IEEEoverridecommandlockouts                              
\usepackage{graphicx}
\usepackage{epsfig} 
\usepackage{mathptmx}
\usepackage{mathtools}
\usepackage{amsmath} 
\usepackage{bm}
\usepackage{cancel}

\usepackage{amssymb}  
\usepackage{amsthm}
\usepackage{siunitx}
\usepackage{wasysym}
\usepackage{wasysym}
\usepackage{xcolor} 
\usepackage[mathcal]{euscript}
\usepackage{bbm}
\usepackage{color}
\usepackage{lipsum}
\usepackage{booktabs}
\usepackage{hyperref}
\hypersetup{colorlinks=true,linkcolor=black,anchorcolor=black,citecolor=black,filecolor=black,menucolor=black,runcolor=black,urlcolor=black}

\usepackage[ruled,vlined,linesnumbered]{algorithm2e}
\usepackage{etoolbox}
\usepackage{bigfoot}
\usepackage{lipsum}
\usepackage{dirtytalk}
\usepackage[normalem]{ulem}

\usepackage{multirow}
\usepackage{bbm}
\usepackage{outlines}
\let\labelindent\relax
\usepackage{enumitem}
\setenumerate[2]{label=\alph*.}
\setenumerate[3]{label=\roman*.}
\usepackage[capitalise]{cleveref}
\crefname{equation}{}{}
\crefname{figure}{Fig.}{Figs.}
\crefname{tabular}{Tab.}{Tabs.}
\crefname{section}{Sec.}{Secs.}

\makeatletter
\patchcmd{\@makecaption}
  {\scshape}
  {}
  {}
  {}
\makeatother

\usepackage[
    style=ieee,
    doi=false,
    isbn=false,
    url=false,
    eprint=false,
    backend=biber,
    natbib=true
    ]{biblatex}

\usepackage{caption}
\bibliography{references}

\usepackage{listings}
\usepackage{tcolorbox}
\tcbuselibrary{listings}
\usepackage{booktabs}

\newtcblisting{mycodelisting}{%
  listing only,
  listing engine=listings,
  colback=gray!5,
  colframe=gray!80,
  boxrule=0.2pt,
  arc=0pt,
  outer arc=0pt,
  top=-5pt,
  bottom=-5pt,
  left=2pt,
  right=2pt,
  fonttitle=\bfseries,
  title=,
  listing options={
    language=Python,
    basicstyle=\ttfamily\footnotesize,
    breaklines=true,
    tabsize=4,
    frame=none
  }
}

\newtheorem*{problem*}{Problem}

\providecommand{\R}{\ensuremath \mathbb{R}}
\providecommand{\N}{\ensuremath \mathbb{N}}
\providecommand{\Z}{\ensuremath \mathbb{Z}}
\newcommand{\NL}{\mathbb{L}}

\newcommand{\regtext}[1]{\mathrm{\textnormal{#1}}}

\newcommand{\tss}[1]{\textsuperscript{#1}}

\newcommand{\lbl}[1]{^{\regtext{#1}}}

\newcommand{\st}{\,\regtext{s.t.}\,}

\DeclareMathOperator*{\argmin}{arg\,min}
\newcommand{\given}{\,\mid\,}
\newcommand{\satisfies}{\models}

\newcommand{\proposition}{\varphi}
\newcommand{\trajectory}{X}
\newcommand{\always}{\Box}
\newcommand{\eventually}{\Diamond}
\newcommand{\taskset}{\Psi}
\newcommand{\task}{\psi}
\newcommand{\schedule}{S}
\newcommand{\gridworld}{W}
\newcommand{\freecells}{F}
\newcommand{\position}{x}
\newcommand{\duration}{\tau}
\newcommand{\msg}{m}
\newcommand{\allrobots}{\mathcal{R}}
\newcommand{\nrobots}{n}
\newcommand{\makespan}{M}
\newcommand{\minsattime}{T}
\newcommand{\confmsg}{\msg\conf}
\newcommand{\confposition}{\position\conf}
\newcommand{\confduration}{\duration\conf}
\newcommand{\allhelpers}{\mathcal{H}}
\newcommand{\capabilities}{c}

\newcommand{\distance}{D}
\DeclareMathOperator{\reason}{NLR}
\newcommand{\prompt}{p}
\DeclareMathOperator{\bnf}{BNF}
\newcommand{\timehorizon}{H}

\newcommand{\glob}{\lbl{g}}
\newcommand{\help}{\lbl{h}}
\newcommand{\conf}{\lbl{c}}
\newcommand{\offr}{\lbl{o}}
\newcommand{\req}{\lbl{r}}
\newcommand{\selection}{\lbl{s}}
\newcommand{\updated}{\lbl{new}}
\newcommand{\opt}{^\star}
\newcommand{\orig}{\lbl{orig}}

\title{\LARGE \bf
Ask, Reason, Assist: Robot Collaboration via \\ Natural Language and Temporal Logic}
\author{
Dan BW Choe\tss{1}, Sundhar Vinodh Sangeetha\tss{2}, Steven Emanuel\tss{3}, \\Chih-Yuan Chiu\tss{1}, Samuel Coogan\tss{1} and Shreyas Kousik\tss{3}
\thanks{
All authors are with the Georgia Institute of Technology, Atlanta, GA, USA.
\tss{1} School of Electrical and Computer Engineering.
\tss{2} School of Aerospace Engineering.
\tss{3} School of Mechanical Engineering.
Corresponding author: \texttt{bchoe7@gatech.edu}
}
}

\begin{document}
\makeatletter
\let\@oldmaketitle\@maketitle
\renewcommand{\@maketitle}{\@oldmaketitle
    \centering
    \captionsetup{type=figure}
    \setcounter{figure}{0}

    \vspace{5pt}
    \includegraphics[width=\linewidth]{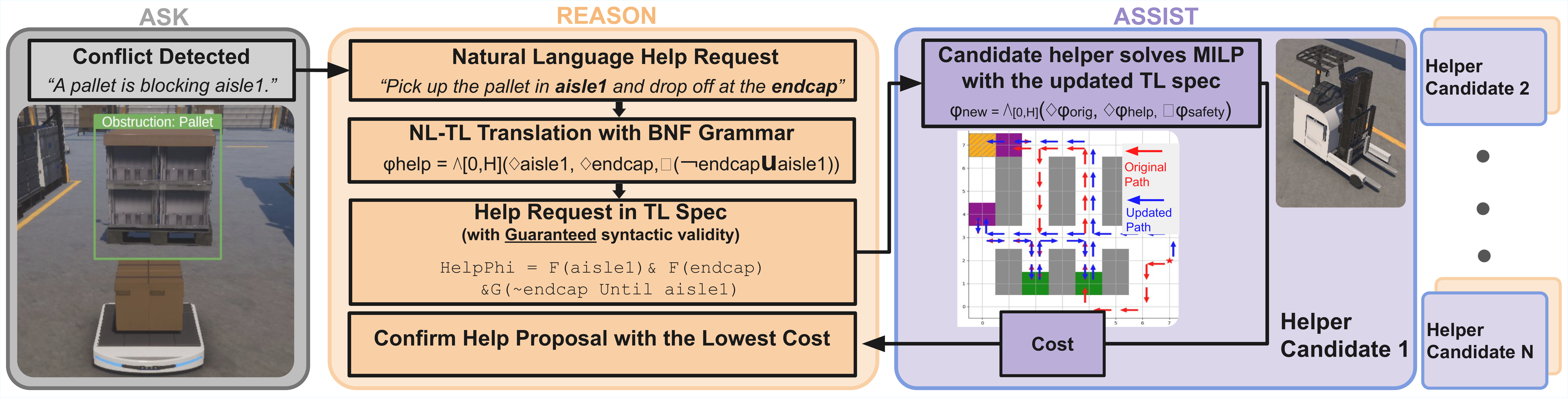}
    \captionof{figure}{\textbf{Overview of the Proposed Framework}: A requester broadcasts a natural language help request, which helpers translate into syntactically valid temporal logic (TL) via a BNF grammar.
    Then each helper independently solves an updated optimal path via MILP to assess the cost associated with the help task and proposes help.
    Finally, the requester selects and confirms the best offer that minimizes the overall impact to the system. 
    Our framework demonstrates how natural language (NL) can serve as a flexible medium for a heterogeneous multi-robot help-request forum.
    Using constrained generation with BNF grammar guarantees valid TL translations, while solving the decentralized MILP optimization problem achieves performance close to a centralized ``Oracle'' baseline.}
    
    \label{fig:overall_framework}
    \vspace{-5pt}
}
\makeatother

\maketitle
\thispagestyle{plain}
\pagestyle{plain} 

\begin{abstract}
Increased robot deployment, such as in warehousing, has revealed a need for collaboration among heterogeneous robot teams to resolve unforeseen conflicts.
To this end, we propose a peer-to-peer coordination protocol that enables robots to request and provide help without a central task allocator.
The process begins when a robot detects a conflict and uses a Large Language Model (LLM) to decide whether external assistance is required.
If so, it crafts and broadcasts a natural language (NL) help request.
Potential helper robots reason over the request and respond with offers of assistance, including information about the effect on their ongoing tasks.
Helper reasoning is implemented via an LLM grounded in Signal Temporal Logic (STL) using a Backus–Naur Form (BNF) grammar, ensuring syntactically valid NL-to-STL translations, which are then solved as a Mixed Integer Linear Program (MILP).
Finally, the requester robot selects a helper by reasoning over the expected increase in system-level total task completion time.
We evaluated our framework through experiments comparing different helper-selection strategies and found that considering multiple offers allows the requester to minimize added makespan.
Our approach significantly outperforms heuristics such as selecting the nearest available candidate helper robot, and achieves performance comparable to a centralized \say{Oracle} baseline but without heavy information demands.

\end{abstract}

\section{Introduction} \label{sec:intro}




Modern warehouses deploy heterogeneous robot fleets, combining mobile robots, forklifts, and manipulators, to meet demanding throughput targets. The diversity of capabilities that makes these teams effective also introduces coordination challenges: robots encounter physical conflicts (e.g., blocked paths) and semantic conflicts (e.g., incompatible task types) that a single robot cannot resolve on its own. At scale, centralized conflict resolution becomes impractical and may require disclosing proprietary schedules. 
Realizing the full potential of these teams requires an efficient peer-to-peer approach where robots request and provide help without revealing private task information.



Recent advances in foundation models, particularly multimodal Large Language Models (LLMs), provides a promising tool in addressing the first step of the conflict resolution process: identifying and describing the conflict \cite{sinha2024real}.
When paired with an embodied agent, these models can generate a concise help request in natural language by processing sensor inputs while leveraging an understanding of robot capabilities \cite{chen2024emos}. Although methods have been presented to use multimodal LLMs for end-to-end conflict resolution in multi-robot settings \cite{kato2024design}, the plans generated by foundation models do not provide the safety guarantees required in environments where humans and robots coexist,nor the spatio-temporal context needed to compute feasible paths under timing constraints. Furthermore, industrial robots frequently operate under strict time constraints, such as in order fulfillment scenarios where shipments must be staged
at outbound docks by designated pickup times. Without
guarantees on the safety and feasibility of conflict
recovery plans, LLMs alone are insufficient for ensuring
reliable multi-agent conflict resolution in time-sensitive
and safety-critical environments.

One path forward to solving this challenge is to apply formal methods, such as the verification of Signal Temporal Logic (STL) and Linear Temporal Logic (LTL) specifications.
STL and LTL tools
can guarantee that a robot's plan satisfies constraints and have been successfully applied to multi-robot coordination and reconfiguration tasks \cite{guo2015multi}\cite{sun2022multi}.
Recent work has also explored their integration with LLMs to enforce safety constraints during task planning \cite{yang2024plug}. 
However, it remains open how to integrate these formal methods with foundation model reasoning for multi-agent coordination. A key insight is that transmitting formal specifications directly between robots is impractical in heterogeneous fleets: each robot has its own atomic propositions grounded in its capabilities; a STL specification composed in one robot's vocabulary may not be interpreted by another robot without a separate integration layer. Natural language bridges this gap, allowing the requester to describe what is needed while each helper determines how to encode it in its own specification. 


\subsection{Related Work}

\subsubsection{Formal Methods for Robot Coordination}

Formal methods, such as Temporal Logic (TL), provide a precise language for specifying robot objectives with temporal and spatial grounding. 
In particular, Signal Temporal Logic (STL) is well-suited to robotics as it reasons over real-valued trajectories with explicit time bounds \cite{belta2019formal}.
Prior work formulates these specifications as mixed-integer programs (MIPs) for certifiable planning \cite{kurtz2022mixed}.
We adopt this approach as a local solver at each robot, and contribute a coordination protocol for robots to exchange help requests and leverage their private solvers without central oversight.
That is, our work differs on \textit{where} and \textit{how} formal reasoning is invoked. 
Once an LLM translates NL to STL on demand, each potential helper independently solves a compact Mixed Integer Linear Program (MILP) that (i) certifies feasibility and (ii) estimates thee marginal cost of adding the help task given its existing, private task schedule. 
This parallelized approach maintains the rigorous guarantee provided by formal methods while avoiding the need for a centralized planner and full disclosure of potentially proprietary information. 

\subsubsection{Natural Language to Temporal Logic}
While temporal logic is a powerful formal language for expressing constraints on task and motion plans, it requires significant effort and expertise to formulate.
As such, a number of works have sought to automatically translate natural language (NL) instructions and specifications into temporal logic (TL) formulas, specifically using LLMs \cite{chen2023nl2tl, fuggitti2023nl2ltl, mavrogiannis2024cook2ltl, wang2025conformalnl2ltl, fang2025enhancing}.
However, they lack guarantees on the accuracy and syntactic validity of generated temporal logic formulas.
Recently, conformal prediction has been used to quantify the uncertainty of LLM outputs, and applied towards an accuracy guarantee for NL to TL \cite{wang2025conformalnl2ltl}.
Specifically, an LLM is prompted to incrementally translate an NL specification, with the LLM queried multiple times for each translation step.
Then, the frequency of each unique response at each step is used as a measure of uncertainty, and triggers asking for help from a human when high uncertainty is detected.
Syntactic validity in this work is addressed by repeatedly sampling from the LLM with the same prompt and filtering out responses which contain invalid operators or atomic propositions.
We note that there is no guarantee that any of the sampled responses from the LLM will be syntactically valid, and that these filtering rules do not fully capture the syntax rules of temporal logic.
We seek to address the problem of generating correct temporal logic syntax in this work.

\subsubsection{Meta-heuristics in Vehicle Routing Problems}
As the centralized baseline for comparison, assigning a new help request within a fleet can be formulated as a dynamic Vehicle Routing Problem (VRP), a long-studied subject within the operations research field  \cite{dantzig1959truck}. Since  exact optimization is intractable at realistic scales, constructive insertion heuristics (e.g sequential/greedy insertion \cite{solomon1987algorithms}) provide a feasible solution promptly and meta-heuristics like Iterated Local Search (ILS) improve the optimality of the initial solution by introducing local moves with perturbations to escape local minima  \cite{lourencco2003iterated}. We implement this centralized hybrid VRP algorithm (insertion + ILS) as an \say{Oracle} baseline that can reshuffle tasks across all robots to minimize the fleet's overall makespan. In contrast to this centralized VRP approach, our framework requires only brief NL messages and scalar cost bids, where robots never disclose their full schedules, therefore minimizing information exchange without sacrificing scalability.

\subsection{Contributions}
We present a framework for heterogeneous warehouse robots, each subject to local task specifications, to request and offer assistance in NL, while ensuring the safety and feasibility of the resulting recovery plan through formal verification.
In particular, we make the following contributions:
\begin{enumerate}
    \item We propose a method to transform NL specifications into TL specifications with a guarantee on syntactic validity (\cref{subsec: method: help request generation,subsec: method: translating proposals to stl}).

    \item We augment LLM agents with spatial and temporal reasoning capabilities, therefore making progress towards realizing guarantees from formal methods (\cref{subsec: method: optimal path via milp}).

    \item We rigorously evaluate our coordination protocol against both simple heuristics and a centralized \say{Oracle} baseline, demonstrating competitive performance with minimal information exchange.  
    (\cref{subsec: experiments: experiment 2}).

\end{enumerate}

\section{Preliminaries and Problem Statement}\label{sec: problem statement}

Consider robots operating in parallel on a variety of tasks; in this work we use warehousing as our canonical example.
We investigate how a system can achieve decentralized conflict resolution while approaching the global efficiency of a centralized solution through peer-to-peer interactions that minimizes the information demand.
Specifically, we ask: 
\textbf{If a robot detects a conflict that it cannot resolve on its own, how can it request and receive help from a fellow robot while ensuring safety and minimizing impact, measured as the added time cost, on the overall system?}

\subsubsection*{Notation}
The reals are $\R$, natural numbers are $\N$, and integers are $\Z$.
To describe functions operating on natural language, we denote $\NL$ as the set of all natural language utterances.
Subscripts are indices and superscripts are labels.
We use $\gets$ to denote the output of NL reasoning.

\subsubsection*{World and Robots}
We consider a multi-robot system (MRS) with $\nrobots$ robots, and $\allrobots = \{0,1,\cdots,\nrobots-1\}$ denoting the set of all robots.
The robots operate on a discrete grid space,$\gridworld \subset \Z^2$ over discrete timesteps  $t = 0,1,\cdots,\timehorizon$,  where $\timehorizon \in \N$ is a finite time horizon. Robot $i$ occupies cell $\position_i(t) \in \gridworld$ at timestep $t$. 
Robot $i$'s trajectory is a sequence of its position over time $t\in[0,\timehorizon]$, denoted by
    $\trajectory_i = \{x_i(t) \}_{t=0}^\timehorizon \subseteq \gridworld$.
We index a trajectory as $\trajectory_i(j) = \position_i(j)$ at a specific timestep and $\trajectory_i(j:k) = \{x_i(t) \}_{t=j}^k$ for a range of positions.
Given trajectory $\trajectory_i$ of duration $\timehorizon$, its Manhattan distance is
    $\distance(\trajectory_i) = \sum_{t=0}^{\timehorizon-1} \Vert \position_i(t+1) - \position_i(t) \Vert_1$.

Each cell is either free space or a static obstacle, a common setup in modeling multi-robot systems \cite{stern2019multi}; we denote all free cells as  $\freecells \subseteq \gridworld$.
Multiple robots can occupy a free cell concurrently ($\position_i(t) = \position_j(t) \in \freecells$ is allowed); we assume local coordination and collision avoidance are handled by lower-level planning and control for which many solutions exist (e.g., \cite{van2008reciprocal,vaskov2019towards,fridovich2021approximate}).
Finally, each robot has an NL representation of its own capabilities $\capabilities_i \in \NL$ (e.g., ``can lift pallets'' or ``max speed $1~\si{m/s}$'').

\subsubsection*{Tasks and Schedules}
Let $\taskset = \{\task_1, \task_2, \dots, \task_K\}$ be the set of $K$ initial tasks to be completed by the system. A task schedule is a partition of these tasks among the robots, denoted by $\schedule = \{\taskset_0, \taskset_1, \dots, \taskset_{n-1}\}$, where $\taskset_i \subset \taskset$ is the set of tasks assigned to robot $i$, $\bigcup_{i \in \allrobots} \taskset_i = \taskset$.
Finally, $\task_i \cap \task_j = \emptyset$ for $i \neq j$, i.e., no two robots are assigned the same task.

For each robot $i$, its assigned tasks $\taskset_i$ are represented as an STL specification $\proposition_i$ and path $\trajectory_i$ for robot $i$.
All robots must also obey a global specification $\proposition\glob$; see App.\ref{appx: global stl spec} for an example including actuation limits and obstacles.
We assume each robot's tasks are feasible, while we certify syntactic validity via constrained generation (\Cref{subsec: method: translating proposals to stl}).

\subsubsection*{Makespan}
Our goal is for robots to minimize time to complete their own tasks plus help tasks, which is the \textit{makespan} of each robot and the overall system.
Consider robot $i$, with initial position $\position_i(0)\in \freecells$ and corresponding STL specification $\proposition_i$.
If the robot's trajectory $\trajectory_i$ satisfies $\proposition_i$, we write $\trajectory_i \satisfies \proposition_i$.
Note, this expression is a shorthand for $(\trajectory_i,0)\satisfies\proposition_i$, meaning we always evaluate the satisfaction of the trajectory starting from $t = 0$.
We minimize makespan by finding $\trajectory_i$ to take the least time to satisfy $\proposition_i$.
To aid in defining the makespan, define the \textit{time to first satisfaction} 
\begin{align}
    \minsattime(\trajectory_i,\proposition_i) = \min_t
        \left\{ t \mid X_i(0:t) \satisfies \proposition_i \right\}.
\end{align}
Note that $\minsattime(\trajectory_i,\proposition_i) = \infty$ if $\trajectory_i \not\satisfies \proposition_i$.
Then we compute the makespan $\makespan(\position_i(0),\proposition_i)\in \N$ as
\begin{align}\label{eq: individual makespans}
    \makespan(\position_i(0),\proposition_i) = \min_{\trajectory_i}
        \left\{
            \minsattime(\trajectory_i,\proposition_i) \mid 
            \trajectory_i(0) = \position_i(0)
        \right\}.
\end{align}
That is, the makespan finds a trajectory that minimizes the time to first satisfaction of $\proposition_i$ while obeying the robot's initial condition.
We implement \Cref{eq: individual makespans} as an MILP solved with Gurobi \cite{gurobi}.
Further details are in \Cref{sec: experiments}.

\subsubsection*{Communications}
We assume robots can send and receive NL messages $\msg \in \NL$ instantaneously and error-free. We use NL rather than transmitting STL specification $\proposition_i$ directly, as valid STL generation requires receiver-side context such as its atomic propositions. 
We denote $\msg_{i\to j}$ as a message from robot $i$ to robot $j$, and a broadcast message as $\msg_{i\to\allrobots}$.

\subsubsection*{Conflicts}
Each robot can detect a conflict (e.g., using a VLM based conflict detector as in \Cref{fig:overall_framework}), which we denote as a tuple $(\confmsg,\confposition,\confduration) \in \NL \times \gridworld\times \N$ containing a natural language description, a location, and a duration required to resolve it.
We assume $\confposition$ is resolved when a helper robot occupies the same cell as the requester for at least $\confduration$ time steps; we leave low-level coordination to future work.

\begin{problem*}
We now formalize the research problem using the above definitions:
Suppose robot $i$ detects a conflict $(\confmsg,\confposition,\confduration)$. 
Without using centralized task assignment, we seek to identify another robot $j\in \allrobots_{-i} = \allrobots \setminus \{i\}$ among the set of candidate helper robots, and plan its motion such that it resolves the conflict while minimizing the increase in the total makespan across all robots.
That is, create $\proposition_j$ such that $\position_j(t) = \confposition$ for all $t' \in [t,t+\confduration]$ while minimizing the sum of individual makespans $\sum _{i\in\allrobots}\makespan(\position_i(0),\proposition_i)$.
\end{problem*}

\section{Proposed Method} \label{sec: method}


In this section, we introduce the main components of our framework, as depicted in Fig. \ref{fig:overall_framework}. Concretely, Sec. \ref{subsec: method: help request generation} describes how a \textit{requester} robot that requires help generates a help request and broadcasts it to the multi-robot system. Each robot in the system then evaluates its capabilities and availability, and any \textit{helper} robot with the potential to assist formulates a help offer in natural language (NL). Then, Sec. \ref{subsec: method: translating proposals to stl} details how each helper translates its help request from NL to STL, while Sec. \ref{subsec: method: optimal path via milp} describes how each helper computes an optimal path  that minimizes total time impact to the system.
Each potential \textit{helper} then replies to the requester in natural language with its help offer and the associated, predicted time impact to the system. Finally, the requester chooses the helper with the lowest impact and affirms its selection via a help confirmation message.

\subsection{Generating Help Requests, Offers, and Confirmations}\label{subsec: method: help request generation}

We implement a multi-robot communication protocol for collaborative conflict resolution using LLMs.
This protocol involves three capabilities: send, receive and broadcast.
We define three types of messages: help requests, help offers, and help confirmations.
See examples in App.\ref{appx:example_request} and \ref{appx:example_proposal}.

Help requests are broadcast messages sent by a robot $i$ facing a conflict after determining that it requires help, by reasoning over the conflict $(\confmsg, \confposition, \confduration)$ and its own capabilities $\capabilities_i$, conditioned on a prompt $\prompt\conf \in \NL$:
\begin{align}
    \msg_{i\to\allrobots_{-i}}\req \gets \reason(\confmsg, \confposition, \confduration, \capabilities_i \given \prompt\conf),
\end{align}
where $\reason$ abstractly represents NL reasoning, implemented with foundation models.
Help requests describe the scene, location, what is required for the conflict to be resolved, and why the robot cannot resolve the conflict.
To ensure conflict location and requester capabilities are included in help requests, we use constrained generation \cite{beurer2023prompting}.

Help offers are sent by potential helpers in response to a help request. 
Each helper $j$ generates a help offer by independently grounding the NL request into its own STL specification by reasoning over the request, its location, capabilities, and its task schedule, which remains private:
\begin{align}
    \msg_{j \to i}\offr \gets \reason(\msg_{i\to\allrobots_{-i}}\req,\position_j,\capabilities_j \given \prompt\help).
\end{align}
We again use constrained generation to ensure help offers include information about capabilities.
In addition, the help offer includes the duration it will take robot $j$ to provide help $\duration\help_j$ (i.e., how long robot $i$ needs to wait), and the additional time to help relative to completing its original tasks $\duration\updated_j$ (i.e., how much robot $j$ is impacted by having to complete the help request).
We compute these durations below in \Cref{subsec: method: optimal path via milp} (see \Cref{eq: help and updated durations}). 
Finally, the requester confirms the lowest-cost help offer:
\begin{align}\label{eq: confirming lowest-cost help offer}
    j\opt =\argmin_{j \in \allrobots_{-i} }\left(\duration\help_j + \duration\updated_j\right).
\end{align}
The selected helper $j\opt$ then receives the message
$\msg_{i\to j\opt}\selection = \regtext{``accept''}$ while all other candidate helpers receive $\msg_{i\to j}\selection = \regtext{``reject''}$ for all $j \neq j\opt$.

\subsection{Translating Help Proposals to STL Specifications}\label{subsec: method: translating proposals to stl}

To translate help proposals from NL to STL,
we draw from recent work using LLMs \cite{chen2023nl2tl, fuggitti2023nl2ltl, mavrogiannis2024cook2ltl}.
We enhance these NL-to-TL methods by defining a Backus-Naur form (BNF) grammar, a notation system for defining formal languages that can be used to constrain the output of LLMs \cite{wang2023grammar}.
Additionally, we finetune the LLM model using LoRA \cite{hu2022lora}.
Specifically, defining a BNF grammar for STL involves defining unary and binary temporal and boolean operators, allowed predicates, and text representations of temporal logic relations.
BNF constrained generation enforces the syntactic validity of generated temporal logic formulas, allowing us to directly feed LLM outputs into an STL solver 
(details on how we encode STL specifications can found in the appendix). 
However, constrained generation has been shown to degrade task performance when grammar constraints are misaligned with the model's tokenization \cite{beurer2024guiding}.
As such, we carefully design the BNF grammar 
(example in App.\ref{appx:bnf_example}) 
to be sufficiently relaxed, for example including optional whitespace and parentheses, and allowing for arbitrary nesting and recursion, while maintaining the guarantee that the output can be processed using standard STL parsers.
Ultimately, we generate an STL specification by reasoning over the help offer message and conflict location, conditioned on a prompt $\prompt\lbl{STL}$ and subject to the BNF grammar:
\begin{align}\label{eq: help offer to stl}
    \proposition_j\help \gets \reason(\msg_{j\to i}\offr,\position\conf \given \prompt\lbl{STL}, \bnf).
\end{align}
Note, enforcing BNF grammar with constrained generation guarantees syntactically valid temporal logic formulas, but not semantic equivalence to the natural language specification.
We evaluate the impact of this limitation in \Cref{exp: nl to tl}.


\subsection{Solving for Optimal Robot Paths}\label{subsec: method: optimal path via milp}


Suppose an initial task schedule $\schedule\orig=\{\Psi_0,...,\Psi_{n-1}\}$ that approximately minimizes total system makespan has been determined.
Each candidate helper $j \in \allrobots_{-i}$ has an original specification $\proposition\orig_j$ and corresponding optimal path
\begin{align}
    \trajectory_j\orig = \{\position_j(t) \given t = 0,\cdots,\makespan(\position_i(0),\proposition\orig_j)\}    
\end{align}
by minimizing Manhattan distance (simplifies computation over Euclidean distance) 
and time to satisfy the task specification (i.e., makespan) while always obeying global specifications:
\begin{subequations}\label{prog: original path milp}
\begin{align}
    \trajectory_j\orig = \argmin_\trajectory
        \quad & \distance(\trajectory) + \minsattime(\trajectory,\proposition\orig) \\
        \st \qquad &X \satisfies \eventually\proposition\orig_j \land \always\proposition\glob.
\end{align}
\end{subequations}
Once the help offer $\msg\offr_{j\to i}$ is translated into $\proposition\help_j$ as in \Cref{eq: help offer to stl}, robot $j$ computes an updated path that enables it to both help and complete the original tasks:
\begin{subequations}\label{prog: updated path milp}
\begin{align}
    \trajectory_j\updated = \argmin_\trajectory
        \quad & \distance(\trajectory) +
            \minsattime(\trajectory,\eventually\proposition\help_j) +
            \minsattime(\trajectory,\proposition\updated_j) \\
        \st\qquad & X \satisfies \proposition\updated \\ &\proposition\updated_j = \eventually\proposition_j\help \land \eventually\proposition\orig_j \land \always\proposition\glob.
\end{align}
\end{subequations}
The term $\minsattime(\trajectory,\eventually\proposition\help_j)$ represents how long the requester must wait for help, whereas the term $\minsattime(\trajectory,\proposition\updated_j)$ represents the helper's time required to complete the help task plus all other tasks.
This encourages the helper to minimize how long the requester must wait.

We formulate \Cref{prog: original path milp} and \Cref{prog: updated path milp} as MILPs by defining predicates in the STL formula as binary variables over discrete time steps, and solve with Gurobi \cite{gurobi} which provides the spatio-temporal grounding that NL help offer $\msg\offr_{j\to i}$ lacks.

Finally, when crafting a help offer, robot $j$ reports the total time it will take to help \textit{and} the additional time to help relative to its original tasks:
\begin{align}\label{eq: help and updated durations}
    \duration\help_j &= \makespan(\position_j(0),\proposition\help_j) \\
    \duration\updated_j &= \makespan(\position_j(0),\proposition\updated_j) - \makespan(\position_j(0),\proposition\orig_j).
\end{align}
Note that, in our implementation, $\duration\help_j$ and $\duration\updated_j$ are computed as a byproduct of solving \Cref{prog: updated path milp}, instead of solving new MILPs to compute these makespans.

Importantly, our method does not alter the original task schedule $\schedule\orig$; only the chosen helper $j$ augments its plan with the help task.
This contrasts with an approach using a centralized Oracle (cf., \cite{solomon1987algorithms,lourencco2003iterated}), which is free to re-allocate any task from any $\task_k$ to any other robot to minimize the total system cost, defined as the sum of the individual makespan,  $\sum _{i\in\allrobots}\makespan(\position_i(0),\proposition_i)$.
This means that our method sacrifices optimality for scalability and minimal information disclosure, as we assess next.






\begin{figure}
    \centering
    \includegraphics[width=\linewidth]{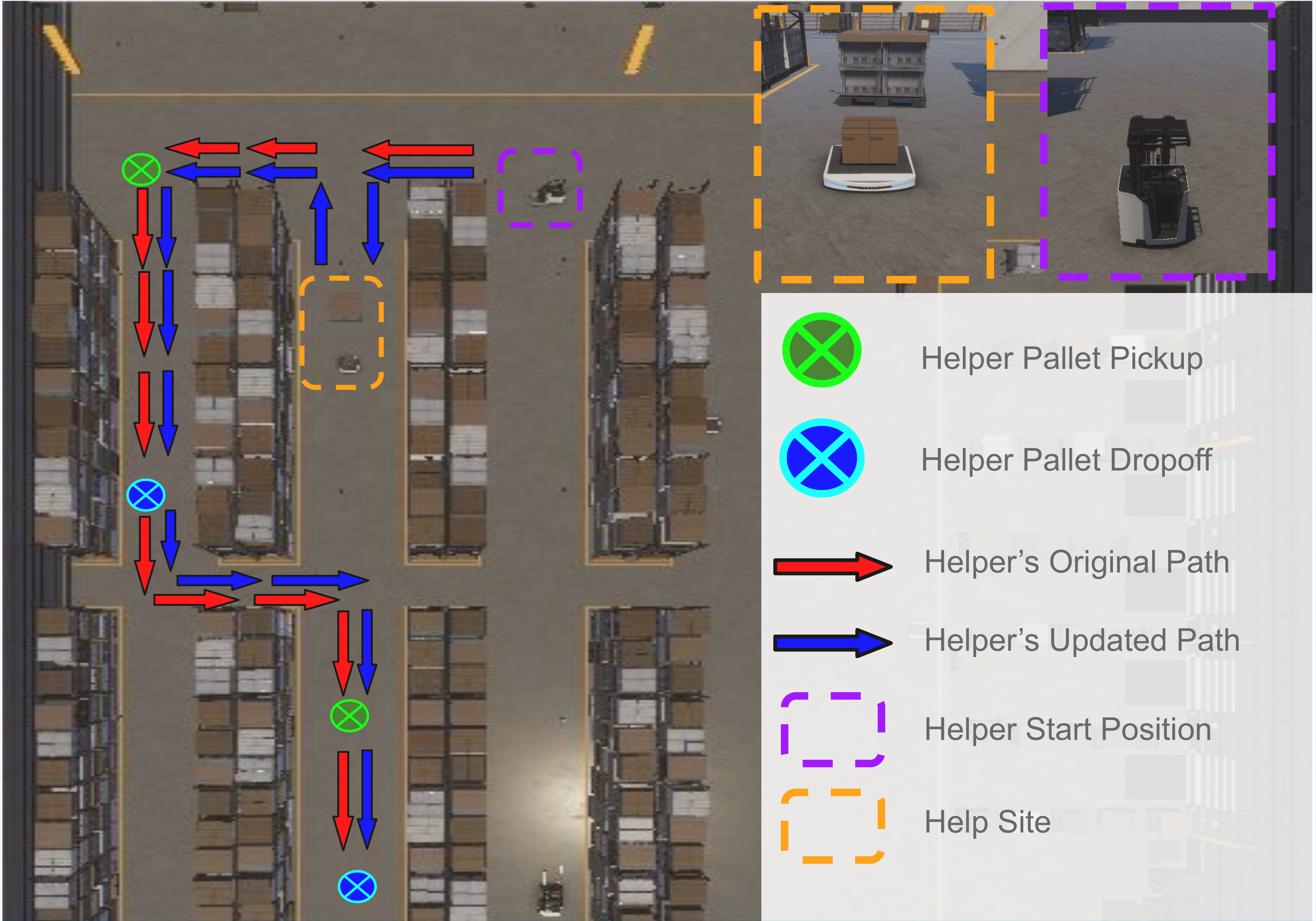}
    \caption{An example of a reconfigured path.
    The help site is reached in 2 time-steps, extending the original path by 2 time-steps for a total cost $\duration\help_j + \duration\updated_j = 4$ time-steps.
}
    \vspace{-20pt}
    \label{fig:path}
\end{figure}
\section{Experiments} \label{sec: experiments}
To study the efficacy of the proposed end-to-end framework for multi-robot collaboration, we isolate and evaluate its two constituent modules: the natural language–to–temporal logic translation component and the temporal logic–to–task plan generation component. Each module is quantitatively compared against state-of-the-art methods in natural language–to–temporal logic translation and vehicle routing, respectively.

All experiments were performed using Python\footnote{Code available \href{https://github.com/ask-reason-assist/ask-reason-assist-codebase}{\textcolor{blue}{\underline{online}} (click for link)}.} 
on a desktop computer with a 32-core i9 CPU, 32 GB RAM, and an NVIDIA RTX 4090 GPU.



\subsection{Experiment 1: Natural Language to Temporal Logic}\label{subsec exp: experiment 1}
\label{exp: nl to tl}

In this experiment, we evaluate our method for transforming natural language (NL) to temporal logic (TL) using BNF constrained generation.
We implement our method with Gemma 3 12B LLM using the \texttt{llama.cpp} library for BNF constrained generation and LLaMA-Factory \cite{zheng2024llamafactory} for finetuning. The BNF grammar is included in the LLM prompt and as a constraint.

\subsubsection*{Hypothesis}
We hypothesize that our method demonstrates comparable translation accuracy to existing baselines on a large and diverse data set of NL-TL pairs, with a strict guarantee on the validity of TL formulas, while running on a significantly smaller LLM (Gemma 3) which can be deployed onboard a robot with a single consumer GPU.

\subsubsection*{Experiment Design}
To benchmark NL to TL translation performance, we use the dataset presented in \cite{chen2023nl2tl}, which consists of 7,500 pairs of natural language sentences and corresponding signal temporal logic (STL) formulas in the context of navigation tasks.
4,500 pairs are used for finetuning, and 3,000 pairs are used for evaluation.

We compare our method to several ablated variants and to a GPT-4 baseline presented in \cite{chen2023nl2tl}.
The ablations we consider include variants without the grammar constraint, without the grammar prompt, and with combinations thereof, allowing us to isolate the contributions of each component of our method.
Our method and each ablation are evaluated with 5 and 20 ``few-shot" examples of natural language tasks with accurate temporal logic translations included in the prompt to the LLM, with 1000 NL-TL pairs randomly sampled from the dataset. Each evaluation is rerun 3 times, with resampled few-shot examples and NL-TL pairs.
In our approach, both the BNF grammar and few shot examples are included in the prompt, and the BNF grammar is enforced during decoding using \texttt{llama.cpp}'s constrained generation.
Results with constrained generation are not presented for GPT-4, as this model can only be accessed through the OpenAI API, which does not allow for a BNF grammar constraint to be specified.

\subsubsection*{Metrics}
We measure Validity as the percentage of generated formulas that are syntactically correct and Accuracy as percentage of valid generated STL formulas logically equivalent to true STL formulas from the dataset.
Accuracy is computed on the mutually valid set of generated formulas across all ablations.
The logical equivalence and syntactic correctness of STL formulas are checked using the \texttt{Spot} library \cite{duret2022spot}.

\subsubsection*{Results and Discussion}
The results are summarized in \Cref{tab:ablation}.
Our method achieves 100\% formula validity in all cases with no loss in accuracy due to constrained generation, confirming our hypothesized guarantee.
Constrained generation results in a modest increase in inference time of 218.2 ms on average.
Additionally, we show improved translation accuracy in the navigation dataset with a significantly smaller LLM.

While semantic accuracy generally improves with advancements in large language models, adhering to the rigid, domain-specific formal language syntax remains a distinct problem.
Additionally, although large models like GPT-4, which has approximately 1.8 trillion parameters compared to our model's 12 billion parameters, can often produce valid syntax without explicit grammar constraints, their computational requirements prevent local deployment.
The proposed method ensures syntactic correctness independent of model capacity, facilitating the use of smaller models for edge deployment.

For each STL formula which was incorrectly translated from natural language by the LLM, we check containment to determine the severity of the failure. In other words, if the true STL formula is contained in the STL formula generated by the LLM, then every TAMP that satisfies the LLM formula will also satisfy the true formula, and therefore the constraint expressed in the natural language will not be violated.
Formally,
\begin{equation}
\proposition\lbl{LLM} \implies \proposition\lbl{true} 
\iff \forall \task \in \taskset: \task \models \varphi\lbl{LLM} \Rightarrow \task \models \varphi\lbl{true}.
\end{equation}

We report the percentage of LLM generated STL formulas which contain the true STL formula in Table \ref{tab:failure_analysis}. 



\begin{table}[ht]
    \centering
    \resizebox{\columnwidth}{!}{%
    \begin{tabular}{clcc}
        \toprule
        \textbf{\# Ex.} & \textbf{Method Variant} & \textbf{Validity (\%)} & \textbf{Accuracy (\%)}\\
        \midrule
        \multirow{4}{*}{5} 
        & Gemma F + P + C (Ours)         & \textbf{100.0} $\pm$ 0.00 & \textbf{99.24} $\pm$ 0.62  \\
        & \phantom{Gemma} F + P          & 99.53 $\pm$ 0.31  & 99.11 $\pm$ 0.82  \\
        & \phantom{Gemma} F              & 93.73 $\pm$ 6.41  & 92.94 $\pm$ 3.88  \\
        & GPT-4 F + P                    & 99.87 $\pm$ 0.12  & 63.83 $\pm$ 8.15  \\
        \midrule
        \multirow{4}{*}{20} 
        & Gemma F + P + C (Ours)         & \textbf{100.0} $\pm$ 0.00  & \textbf{98.44} $\pm$ 0.1  \\
        & \phantom{Gemma} F + P          & 99.40 $\pm$ 0.53  & 98.03 $\pm$ 0.26  \\
        & \phantom{Gemma} F              & 99.20 $\pm$ 0.40  & 93.54 $\pm$ 1.97  \\
        & GPT-4 F + P                    & 99.87 $\pm$ 0.23  & 93.81 $\pm$ 2.04  \\
        \bottomrule
    \end{tabular}
    }
        \caption{Ablation study on NL to TL with varying number of few-shot examples. F, P, and C refer to few shot prompting, inclusion of the BNF grammar in the prompt, and the BNF grammar constrained generation respectively.}
        \vspace{-15pt}
    \label{tab:ablation}        
\end{table}

\begin{table}[ht]
    \centering
    \resizebox{0.8\columnwidth}{!}{
    \begin{tabular}{clc}
        \toprule
        \textbf{\# Ex.} & \textbf{Method} & \textbf{LLM $\implies$ True (\%)} \\
        \midrule
        \multirow{4}{*}{5}  
        & Gemma F + P + C (Ours)  & 99.73 $\pm$ 0.47 \\
        & \phantom{Gemma} F + P   & 99.73 $\pm$ 0.31 \\
        & \phantom{Gemma} F       & 97.64 $\pm$ 1.68 \\
        & GPT-4 F + P & 75.15 $\pm$ 12.15\\
        \midrule
        \multirow{4}{*}{20} 
        & Gemma F + P + C (Ours)  & 99.05 $\pm$ 0.43 \\
        & \phantom{Gemma} F + P   & 98.77 $\pm$ 0.74 \\
        & \phantom{Gemma} F       & 96.12 $\pm$ 2.01 \\
        & GPT-4 F + P & 96.46 $\pm$ 2.33\\
        \bottomrule
    \end{tabular}
    }
    \caption{Containment checking of LLM generated STL formulas for each NL to TL ablation.
    If the true STL formula is contained in the LLM-generated STL formula, the constraint from natural language will always be satisfied.}
    \label{tab:failure_analysis}
    \vspace*{-10pt}
\end{table}


\subsection{Experiment 2: Mobile Robot Blocked by a Pallet}
\label{subsec: experiments: experiment 2}


In our second experiment, we compare our decentralized framework to a centralized ``Oracle" in a scenario where forklift robots respond to a help request from a mobile robot prompted by the scene description $\confmsg =$ ``A pallet is blocking the entrance to the picking aisle.''
Each forklift robot $j \in \allhelpers$ then evaluates whether it can handle this additional task (moving the obstructing pallet) on top of its existing pick-and-place (PNP)\footnote{Our STL specification of a PNP job is in App.\ref{appx: pick and place specification}.} jobs by updating its STL specification $\proposition\orig_j$. 
The help task, $\proposition_j\help$, is similarly encoded as a PNP job where the forklift must travel to the help site, pick the obstructing pallet, and place it in the nearest free cell.

\subsubsection*{Hypothesis}
We hypothesize that our decentralized framework, where each potential helper solves for its updated TL specification via MILP in parallel, will achieve system efficiency comparable to a near-optimal ``Oracle'' with full system knowledge.
We expect our method to significantly outperform myopic, distance-based heuristics, even when such heuristics are coupled with a central optimizer.

\subsubsection*{Experiment Design and Baseline}

We conducted 100 simulation trials with randomly chosen help-site locations, with a single starting scenario to ensure fair comparison within each trial. 
We spawn six forklift robots at random initial positions and distribute twelve PNP tasks among them per a global schedule, $\schedule\orig$.
We pre-compute the schedule using the \say{Oracle} baseline to find a near-optimal solution.
When a help task $\proposition\help$ arises, the Oracle finds a new schedule via Iterated Local Search (ILS) metaheuristics, as detailed\footnote{The algorithm table can be found in App.\ref{app:oracle}} in \cite{lourencco2003iterated}. 
Our method and the three distinct baselines were evaluated against this identical initial setup: (B1) The ILS-based global planner \say{Oracle}, (B2) the closest forklift to the help site solves for the optimal path via MILP, and (B3) a hybrid approach where the help task is assigned to the closest agent, but the Oracle re-optimizes all other tasks.

\subsubsection*{Results and Discussion}

\begin{figure}
    \centering
    \includegraphics[width=0.8\columnwidth]{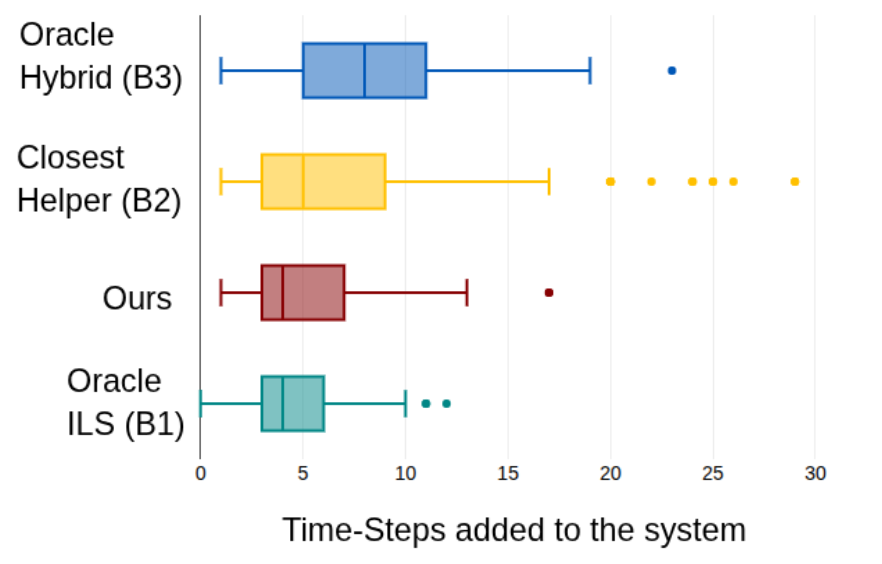}
    \caption{Box plot comparison of total time-steps added to the system under different methods tested in \cref{subsec: experiments: experiment 2}. Our method tracks the Oracle solution within 22\% (mean) while significantly outperforming the distance based heuristics (B2) and hybrid approach (B3). 
}
    
    \vspace{-15pt}
    \label{fig:box}
\end{figure}

\Cref{fig:box} summarizes our findings.
The Centralized Oracle (B1) added an average of 4.49 time-steps to the system by inserting the help task, whereas our method achieved a mean of 5.47 time-steps, within 18\% of the centralized baseline. This demonstrates that informed local optimization captures most of the performance gains without global task reassignment or full information disclosure. Importantly, the average \texttt{Gurobi} model solve time was 5.3 seconds with 30 discretized time horizon, demonstrating computational practicality for real-time deployment.

On the other hand, our method which selects the helper offering the lowest \emph{total cost (\textbf{$\duration\help_j + \duration\updated_j$})} provides 46\% and 53\% efficiency gains over B2 and B3 respectively.
We observe that the inter-dependencies in PNP jobs make distance-based heuristics (B2-B3) suboptimal, consistent with previous results in operations research \cite{de2007design}.
Furthermore, an initial myopic decision (B3) can constrain the overall system's optimality from which even a powerful optimizer cannot recover efficiently \cite{cormen2022introduction}. In contrast, our method explicitly avoids such myopia by reasoning over future task interactions.

\section{Demonstrations}
\begin{figure*}

    \centering
    \includegraphics[width=\linewidth]{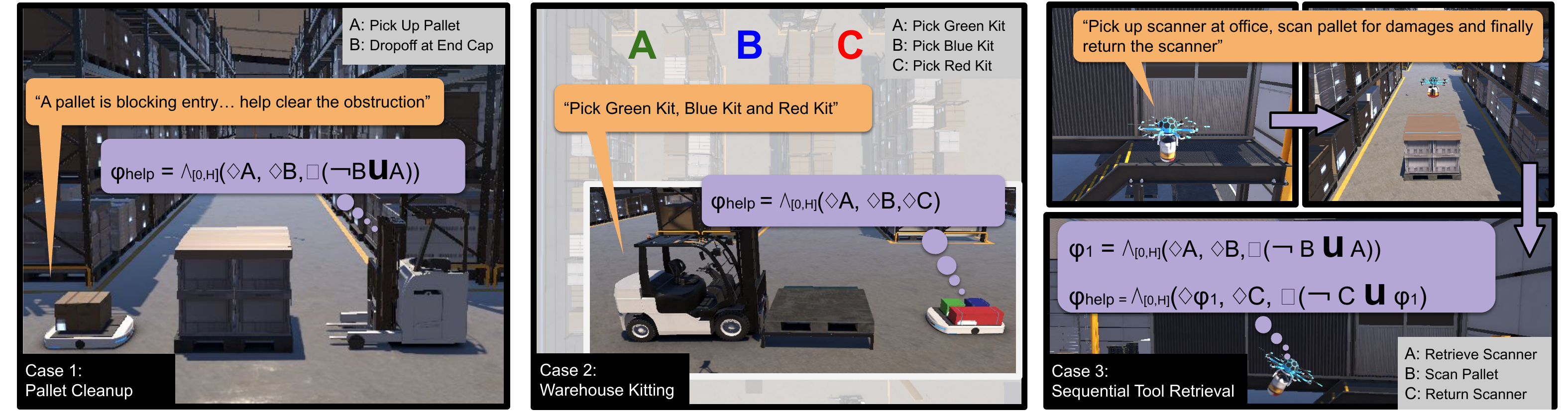}
    \caption{Unity-based demonstrations of our NL-to-TL decentralized framework. (Left) \textbf{Pallet Cleanup}: a forklift responds to an NL request to clear a blocked aisle, updating its MILP plan to integrate the help task. (Center) \textbf{Warehouse Kitting}: unordered conjunctive goals (``pick A,B,C'') are translated into a concise TL formula allowing flexible task sequencing. (Right) \textbf{Sequential Tool Retrieval}: strict temporal ordering (``first-then-finally'') is captured by the nested TL specification, enabling execution of complex multi-step tasks. These demonstrations showcase our end-to-end framework---where each helper robot can translate natural language into valid temporal logic and update its MILP plan---running in a realistic Unity warehouse simulator.
    }
    \vspace{-10
    pt}
    \label{fig:case studies}
\end{figure*}

We now demonstrate the utility of NL in conflict resolution on three complex warehouse tasks.
These complement our experiments, which stress-test components of our method.
Here, we illustrate our framework's practicality in a high-fidelity physics-based Unity simulator.
In particular, we integrate our NL-to-TL translation module with a MILP-based TAMP mechanism for executing sequentially nested tasks, alongside local collision avoidance and an A* path planner for realistic motion plans.
Full videos of each demo are available on our \underline{\href{https://ask-reason-assist.github.io}{\textit{project website}}}.

\subsubsection*{Demo 1: Pallet Cleanup}
First, we revisit the pallet-blocking task from the second experiment, where forklift robots must help a mobile robot.
The NL help request, ``A pallet is blocking the aisle'', is translated into a syntactically valid TL specification for the MILP solver based TAMP module updates the optimal forklift agent's task schedule.
This demo showcases a complete rollout of the framework to complement the experiment in \Cref{subsec: experiments: experiment 2}.

\subsubsection*{Demo 2: Warehouse Kitting}
Our second scenario is a warehouse kitting task, common in fulfillment operations. Here the NL input is, \say{Eventually pick up an item from Aisle A (Green Kit) and Aisle B (Blue Kit) and Aisle C (Red Kit)}.
Our NL-TL module successfully translates this into a semantically correct and syntactically valid TL formula:
\begin{align}
    \proposition\help = \bigwedge_{[0,\timehorizon]}\left(\eventually \position\lbl{pick A}, \eventually \position\lbl{pick B}, \eventually \position\lbl{pick C}
    \right)
\end{align}
Our system uses this formula to compute the optimal sequence of visits to minimize the total makespan added to the system. 
This demo showcases our system's ability to handle unordered conjunctive goals (i.e., tasks that can be done in any order), as the LLM correctly translates the semantics that the first three APs can become true in any order.

\subsubsection*{Demo 3: Sequential Tool Retrieval}
In our final, most complex scenario, the request is, ``First visit the front office to pick up the high fidelity scanner, then travel to the help site to scan the damaged good, and finally navigate back to the front office to return the scanner,'' 
translated to STL as
\begin{align*}
    &\proposition^1= \bigwedge_{[0,\timehorizon]}\left(\eventually \position\lbl{pickup Scanner}, \eventually\position\lbl{scan}, \always(\lnot \position\lbl{scan} \textbf{U} \position\lbl{pickup Scanner}) \right)\\
    &\proposition\help = \bigwedge_{[0,\timehorizon]}\left(\eventually\proposition^1,\eventually\position\lbl{return Scanner},\always(\lnot \position\lbl{return Scanner}\textbf{U} \proposition^1)\right)
\end{align*}
Our NL-to-TL translation module correctly interprets the keywords "First, then and finally'' to enforce a strict order in which each AP holds true, highlighting how our framework enables a robot to autonomously leverage the expressiveness of complex TL specifications. 
This demo showcases the framework's practicality for capturing strict sequential and nested temporal dependencies from natural language.
\section{Conclusion and Future Work}\label{sec: conclusion}

This paper presents a novel framework for robots to request and receive help for decentralized conflict resolution.
Our method lets robots coordinate and reason over their own capabilities in natural language (NL) while reasoning over their motion plans and tasks in formal temporal logic (TL).
We achieve this via a method for converting between NL and TL with guaranteed validity, thereby giving LLM agents spatial and temporal reasoning.
Through experimental evaluation, we find that our method can resolve conflicts while maintaining overall system performance similar to a centralized Oracle baseline.
Furthermore, demonstrations across several robot morphologies and capabilities show how our method successfully combines the flexibility of NL with the formality of TL.
That said, our method still has limitations to address in future work.
First, we assume conflicts are detected correctly.
Second, we have no considered low-level motion planning or physics in helper or requester capabilities.
Third, we have only used NL in a specific help request problem setting, but it remains open how NL can integrate into broader multirobot operations.







\renewcommand{\bibfont}{\normalfont\footnotesize}
{\renewcommand{\markboth}[2]{}
\printbibliography}

\begin{appendices}
\crefalias{section}{appendix}
\crefalias{subsection}{appendix}
\section*{Appendix}\label{sec:appendix}

\subsection{Encoding STL Specifications}\label{appx: encoding stl}
We propose a lightweight interface to express navigational tasks as STL specifications.
Inspired by \texttt{stlpy} \cite{kurtz2022mixed}, our method emphasizes extracting \emph{spatial} propositions and decoupling \emph{temporal} constraints.
We provide an illustrative example here.
Consider encoding ``\textit{Visit Aisle 1 at least once before time $T$ while avoiding obstacles},'' in STL:
\begin{align}
    \proposition\lbl{spec} = \always_{[0,\timehorizon]} \lnot \: \regtext{Obstacle} \land \eventually _{[0,T]}\: \regtext{Aisle1}
    \label{eq: stlpy vs our stl example}
\end{align}
where Aisle1 and Obstacle are atomic propositions (AP) tied to the corresponding world coordinates (true when the agent is in the region).
We specify the time horizon as
 \begin{mycodelisting}
model.spec = F(aisle1) & G(NOT(obstacle))
model.T = T
\end{mycodelisting}
\subsubsection{Global STL Specification}\label{appx: global stl spec}

Each robot is subject to global safety and actuation constraints.
For example, all ground vehicles must avoid obstacle cells and can only move one grid space in any cardinal direction):
\begin{equation}
\proposition\lbl{global} = \always(\lnot \regtext{Obstacle} \land \mathcal{L}\lbl{actuation})
\end{equation}
\vspace{-20pt}
\subsubsection{Pick-and-Place (PNP) Tasks as STL Specifications}\label{appx: pick and place specification}
\begin{align*}
    &\varphi_{1} = (\lnot \position\lbl{place} \; \textbf{U}_{[0,\timehorizon]} \:\position\lbl{pick})\\
    &\textit{(Robot must pickup the pallet before placing it)}\\
    &\varphi_{2} = \Box_{[0,\timehorizon]}(\position\lbl{pick}\rightarrow(\lnot \position\lbl{others} \textbf{U}_{[0,\timehorizon]}\position\lbl{place})\\
    &\textit{(Pallet picked up must be placed before starting other tasks)}\\
    &\varphi_{3} = \Diamond_{[0,\timehorizon]} \position\lbl{place}\\
    &\textit{(Pallet is eventually placed within time horizon $\timehorizon$)}\\
    &\varphi_{\text{pnp}}(\position\lbl{place}, \position\lbl{pick}) = \varphi_1 \land \varphi_2 \land \varphi_3 
\end{align*}
\subsubsection{Example STL BNF Grammar}\label{appx:bnf_example}
$ $
\begin{mycodelisting}
root ::= ws expr ws
expr ::= term (binary-op term)*

term ::= atomic-formula | unary-op ws "(" ws expr ws ")" | unary-op ws atomic-formula | "~" ws term | "(" ws expr ws ")"

predicate-name ::= "go_to_charger" | "go_to_rack_A"
atomic-formula ::= predicate-name | "(" ws predicate-name ws ")"
ws ::= [ \t\n]*

binary-op ::= ws ("&" | "|" | "->" | "U") ws
# '&' (and): both propositions must be true
# '|' (or): at least one predicate must be true
# '->' (implies): if A is true, then B must be true
# 'U' (until): A must be true until B is true

unary-op ::= "G" | "F"
# G (globally): Predicate must always be true at every timestep
# F (eventually): Predicate must be true at some time in the future
\end{mycodelisting}

\subsection{Natural Language}
\subsubsection{Example Help Request}\label{appx:example_request}

\begin{quote}
\sffamily
\footnotesize
Mobile Robot ID 1: \\
``A pallet is blocking the aisle at location (1, 4). Assistance is required to move the pallet. Pick up pallet at (1,4) and drop it off at the closest free drop zone."

\end{quote}

\subsubsection{Example Help Offer}\label{appx:example_proposal}
\begin{quote}
\sffamily
\footnotesize
Forklift ID 3:\\
``I can help you in 5 minutes, but it will add 8 minutes to my overall makespan."


\end{quote}

\subsection{Oracle Baseline with the ILS Algorithm}\label{app:oracle}

\begin{table}[h!]
\vspace{-3mm}
\centering
\label{alg:oracle}
\begin{tabular}{l l}
\hline
\multicolumn{2}{l}{\textbf{Algorithm:} Oracle ILS} \\
\hline
\multicolumn{2}{l}{\textbf{Input:} Cost function $J$, Neighborhood $\mathcal{N}$} \\
\multicolumn{2}{l}{\textbf{Output:} Near-optimal schedule $S^*$} \\
1: & $S_0,S, S^* \leftarrow \text{GreedyInsertion}(J)$ \\
2: & \textbf{for} $k=1$ \textbf{to} max\_iterations \textbf{do:} \\
3: & \quad $S\lbl{cand} \leftarrow \text{Perturb}(S)$ \textit{// Perturbation Step} \\
4: & \quad $S' \leftarrow \text{LocalSearch}(S\lbl{cand}, \mathcal{N})$ \textit{// Local Search Step} \\
5: & \quad \textbf{if} $J(S') < J(S)$ or $\text{rand()} < P(S', S)$ \textbf{then:} \\
6: & \qquad $S \leftarrow S'$ \textit{// Acceptance Criterion} \\
7: & \quad \textbf{if} $J(S) < J(S^*)$ \textbf{then:} \\
8: & \qquad $S^* \leftarrow S$ \\
9: & \textbf{return} $S^*$ \\
\hline
\end{tabular}
\end{table}
\end{appendices}
\end{document}